\documentclass[twoside,11pt]{article}

\usepackage[numbers,sort]{natbib}
\usepackage[abbrvbib]{jmlr2e}

\usepackage{amsmath}
\usepackage{mathtools}
\usepackage[ruled,vlined,linesnumbered]{algorithm2e}
\usepackage{algorithmic}
\usepackage{subcaption}
\usepackage{booktabs}

\newcommand*{\ImgPath}{img/}

\newcommand*{\ExpPath}{exp-results/}

\newcommand{\cmt}[1]{}
\newcommand{\todo}[1]{}
\newcommand{\later}[1]{}

\newcommand{\refsec}[1]{Section~\ref{sec:#1}}
\newcommand{\labelsec}[1]{\label{sec:#1}}
\newcommand{\reffig}[1]{Figure~\ref{fig:#1}}
\newcommand{\labelfig}[1]{\label{fig:#1}}

\newcommand{\name}{\textsc{Rome}}
\newcommand{\namep}{\textsc{Rome}~}

\newcommand{\specialcell}[2][c]{%
  \begin{tabular}[#1]{@{}c@{}}#2\end{tabular}}

\newcommand{\E}{\mathbb{E}}
\newcommand{\Var}{\mathrm{Var}}
\newcommand{\g}{\;|\;}

\usepackage{scalerel,stackengine,amsmath}

\newcommand{\V}{\mathrm{Var}}
\newcommand{\Cov}{\mathrm{Cov}}
\newcommand{\MSE}{\mathrm{MSE}}

\newcommand{\D}{\mathcal{D}}
\newcommand{\Beta}{\mathrm{Beta}}

\newcommand{\IG}{\mathrm{I}}
\newcommand{\Hh}{\mathrm{H}}
\newcommand{\KL}[2]{\mathrm{KL}\left[ #1 \;||\; #2 \right]}

\newcommand{\eq}[1]{Eq.~\ref{eq:#1}}
\newcommand{\labeleq}[1]{\label{eq:#1}}

\newcommand{\nnnl}{\nonumber \\}

\newcommand{\defeq}{\vcentcolon=}

\begin{document}

\title{Residual Overfit Method of Exploration}

\author{\name James McInerney \email jmcinerney@netflix.com \\
       \addr Netflix
       \AND
       \name Nathan Kallus \email kallus@cornell.edu \\
       \addr Cornell University \& Netflix 
}

\editor{}

\maketitle

\begin{abstract}

Exploration is a crucial aspect of bandit and reinforcement learning algorithms. The uncertainty quantification necessary for exploration often comes from either closed-form expressions based on simple models or resampling and posterior approximations that are computationally intensive. We propose instead an approximate exploration methodology based on fitting only two point estimates, one tuned and one overfit. The approach, which we term the residual overfit method of exploration (\name), drives exploration towards actions where the overfit model exhibits the most overfitting compared to the tuned model. The intuition is that overfitting occurs the most at actions and contexts with insufficient data to form accurate predictions of the reward. We justify this intuition formally from both a frequentist and a Bayesian information theoretic perspective. The result is a method that generalizes to a wide variety of models and avoids the computational overhead of resampling or posterior approximations. We compare \namep against a set of established contextual bandit methods on three datasets and find it to be one of the best performing.

\end{abstract}

\section{Introduction}

\cmt{bandits and RL}
\cmt{interaction is the new norm in machine learning}

The use of machine learning in interactive environments such 
as recommender systems~\citep{li2010contextual, vanchinathan2014explore, mcinerney2018explore} and display ads~\citep{li2010exploitation, chapelle2011empirical, jeunen2019learning} 
motivates the study of 
how to balance taking high value actions (exploitation) 
with gathering diverse data to learn better models (exploration). 
The framework of contextual multi-armed bandits,  
and its extension to reinforcement learning in dynamic environments,  
provides guidance for addressing this important task \citep{sutton2018reinforcement}. 
Generally, efficient algorithms tackle the trade-off by encouraging 
actions with high model uncertainty, 
either by adding an explicit bonus for uncertainty as in upper confidence bound (UCB) algorithms \citep{lai1985asymptotically} or by sampling from the posterior distribution over the parameters to promote uncertain actions as in Thompson sampling 
\citep{chapelle2011empirical}. In either case, some quantification of uncertainty is needed.

There are several challenges to uncertainty quantification for both UCB and Thompson-sampling algorithms.
With the exception of simple models like linear and/or conjugate priors, either the sampling or posterior distributions are not analytically known, and instead approximation methods are needed, 
such as bootstrapping, Markov chain Monte Carlo (MCMC), or variational inference (VI) \citep{bishop2006pattern}. 
Both bootstrapping and MCMC are computationally intensive 
and the latter requires diagnostics to assess convergence. 
VI is scalable but tends to require a specialized algorithm for each class of model to be effective 
and has the property of underestimating posterior variance due 
to its objective being an expectation with respect to the approximating distribution \citep{mackay1992information}.

\cmt{ideally use posterior, a number of computational hurdles}
\cmt{Bayesian exploration: based on parameter uncertainty, not just prediction uncertainty}
\cmt{Bayesian uncertainty gives this in principle but expensive to achieve in practice (computational, restructuring existing systems)}
\cmt{practical hurdles: need to reimplement all models, hard to adapt existing model for exploration, other than very simple epsilon greedy or boltzmann}
\cmt{modes of interaction: separate modeling from exploration or solve both together? separation of concerns.}
\cmt{need for embarrassingly easy to implement exploration} 

Against this background, 
our motivation is to develop 
an effective methodology for exploration 
that is scalable and adaptable to a wide range of models. 
Crucially, we seek a method of uncertainty quantification that applies to complex predictive models that may be biased.
Bias is introduced because 
the best estimators in terms of mean-squared error
use tools to prevent overfitting, such as 
L1/L2 regularization, bagging \citep{breiman1996bagging}, dropout \citep{baldi2013understanding}, early stopping \citep{prechelt1998early, caruana2001overfitting}, and the like. 
While improving prediction, these make uncertainty quantification more difficult as the uncertainty consists of more than just variance.
We show how fitting one additional model 
without these tools, 
i.e., formulating an overfit model of the data, 
and combining its predictions with those of the tuned estimator 
enables approximate uncertainty quantification. 
We formalize this 
as the \emph{residual overfit} 
and we use it to drive exploration in what we call the residual overfit method of exploration~(\name). 

From a frequentist perspective, the residual overfit provides an upper approximation of pointwise uncertainty. From a Bayesian information theoretical perspective, the residual overfit provides an upper approximation of the information gain from exploring at any one new point.
These two perspectives suggest it is a possible approximation for driving exploration. To explore this in practice, we consider a bandit experimental setup and compare both UCB and Thompson sampling algorithms based on \namep to benchmark methods that either use resampling to tackle complex models or use exact uncertainty quantification for simple models. Across our experiments, we find \namep performs competitively, often getting the best performance, despite its simplicity and tractability. 
Together, our results suggest \namep is a good option for driving exploration in practical settings with complex predictive models.

\section{The Residual Overfit} 
\labelsec{residual_overfit_1}

\begin{figure}
\centering

\begin{subfigure}[b]{.49\linewidth}
\includegraphics[width=\linewidth]{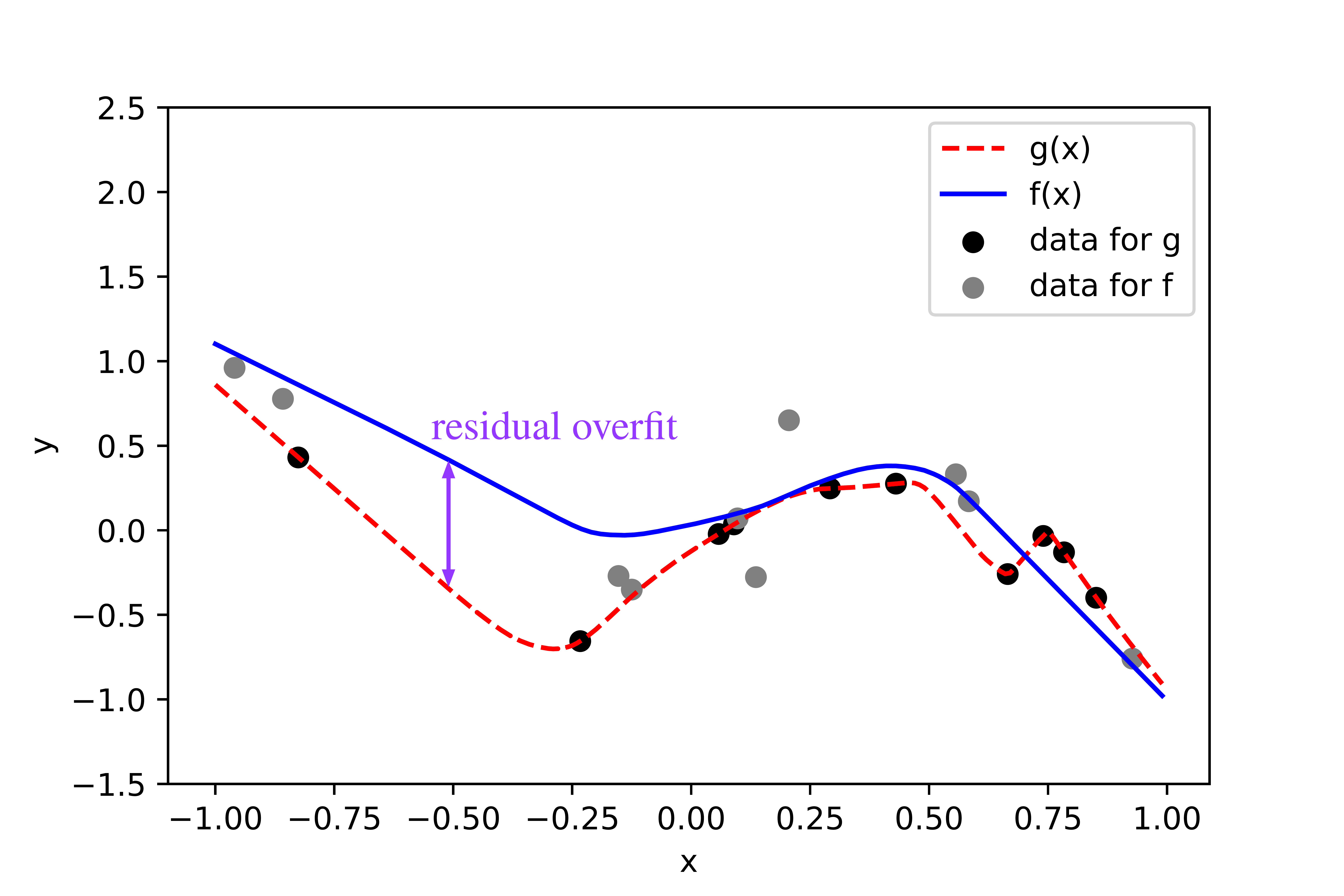}
\caption{Example of fitting regularized model $f$ (blue solid line) and overfit model $g$ (red dashed line) to toy data. 
		The absolute difference between the two functions at any $x$ is the residual overfit. 
		To ensure independence between the fits, they are made on 
		an equal-sized random split of the 20 observations.}
\labelfig{abc}
\end{subfigure}
\hfill
\begin{subfigure}[b]{.49\linewidth}
\includegraphics[width=\linewidth]{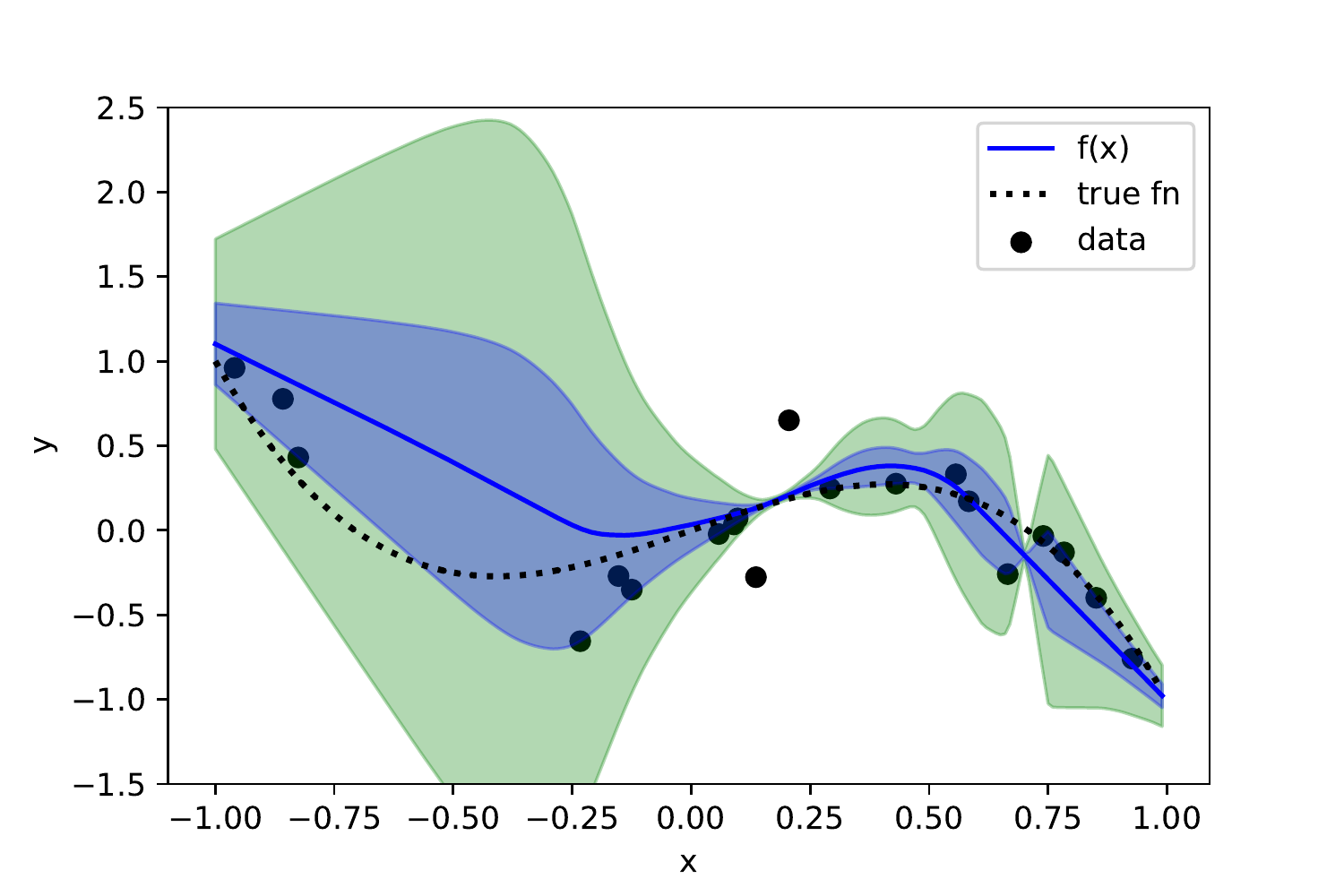}
\caption{Using the same data and fits as \reffig{abc}, 
the $f$ function is annotated with the residual overfit (inner blue shaded area) 
and 2.58 times the residual overfit representing 
an approximate 99\%~confidence interval (outer green shaded area). 
The mean of the true data generating distribution is 
plotted in black.}
\labelfig{rome_eb_toy}
\end{subfigure}

\end{figure}

\paragraph{Definitions}
Given a design $x_1,\dots,x_n$, we consider data consisting of noisy observations of a function $h$, $y_i = h(x_i) + \epsilon_i$, where $\epsilon_i$ has mean zero and variance $\sigma^2(x_i)$ and is independent of $\epsilon_j$ for $j\neq i$.
Let $\D=(x_1,y_1,\dots,x_n,y_n)$ represent the observed data.
Considering the design fixed, the randomness of $\D$ reduces to the randomness of $\epsilon_1,\dots,\epsilon_n$. The design may also be random and all the conclusions would still follow since they hold for any one design.
Let $f$ and $g$ be estimators for the unknown function $h$ based on the training data $\D$. 
Function $f$ is generally understood to be trained for estimation, i.e., for the lowest mean squared error ($\MSE$).
Function $g$ is trained for unbiasedness, i.e., it satisfies the constraint $\E[g(x)] = h(x)$ for any $x$, where expectations are taken with respect to the training data, that is, with respect to the randomness of $\epsilon_1,\dots,\epsilon_n$. 
We define the {\bf residual overfit} at $x$ as,
\begin{flalign}
	s(x) &= | f(x) - g(x) |. \labeleq{residual_overfit}
\end{flalign}

{ \proposition 
When $f$ and $g$ are independent, 
the expected squared residual overfit at $x$ is equal to 
the mean squared error of $f$ plus the variance of $g$,
\begin{flalign}
	\E[s(x)^2] = \MSE[f(x)] + \Var[g(x)],  \labeleq{residual_overfit_summary}
\end{flalign}
where $\MSE[f(x)]=\E[(f(x)-h(x))^2]$.
Recall $\E$ and $\Var$ are taken with respect to the training data.
Note $x$ is a fixed input and is not random.
}

\vspace{10pt}

\begin{proof} We have
\begin{flalign}
\E[s(x)^2]&=\E[((f(x)-h(x))-(g(x)-h(x)))^2]\\
&=\MSE[f(x)]+\MSE[g(x)]+\Cov(f(x)-h(x),g(x)-h(x))\\
&=\MSE[f(x)] + \Var[g(x)],
\end{flalign}
where the last equality is by the unbiasedness of $g$ and the independence of $f$ and $g$.
\end{proof}

The estimators $f$ and $g$ can be made independent by training on two disjoint random splits of the data.

\todo{extend discussion to $f$ being low variance high bias, i.e. well regularized, and $g$ is high variance low bias, i.e. overfit. 
this makes $\MSE[g(x)] = \Var[g(x)]$, this is good because it means that the only bias in $s(x)^2$ comes from $f$ which is what we want.}

\todo{discuss 1D toy example}

\paragraph{Why not fit a model to the prediction error of $f$ instead?} 
An appealing alternative in studying the error of $f$ may be to directly fit a predictive model of the squared error of $f$. This, however, involves \emph{both} the estimation error and the noise. Namely, fix $x_0$ and consider $y_0 = h(x_0) + \epsilon_0$; then predicting the error of $f$ at $x_0$ would estimate $\E[(f(x_0)-y_0)^2]=(f(x)-h(x))^2+\sigma^2(x_0)$. Taking expectations over the data as well yields $\E[(f(x_0)-y_0)^2]=\MSE[f(x)]+\sigma^2(x_0)$. Thus, at best, a model for the prediction error of $f$ would involve the irreducible variance $\sigma^2(x_0)$, which does not vanish. Therefore, this may not well reflect the estimation error of $f$. This is known as the white noise problem in reinforcement learning~\citep{schmidhuber2010formal}. 
In contrast, the expected squared residual overfit only involves estimation errors. 
In addition, if the data is heteroskedastic, 
then $\sigma^2(x)$ will vary across $x$ even under no model uncertainty and will incorrectly bias exploration. 

See \reffig{rmse_eb_toy} for a visual comparison of the two approaches.

\cmt{
\vspace{10pt}
figure to show the average uncertainty approximation 
averaged \namep error bars across data runs, 
not as bad as it seems to have any particular 
plot look ``spikey'' or with cross overs between $f$ and $g$
}

\begin{figure}
\centering

\begin{subfigure}[b]{.49\linewidth}
\includegraphics[width=\linewidth]{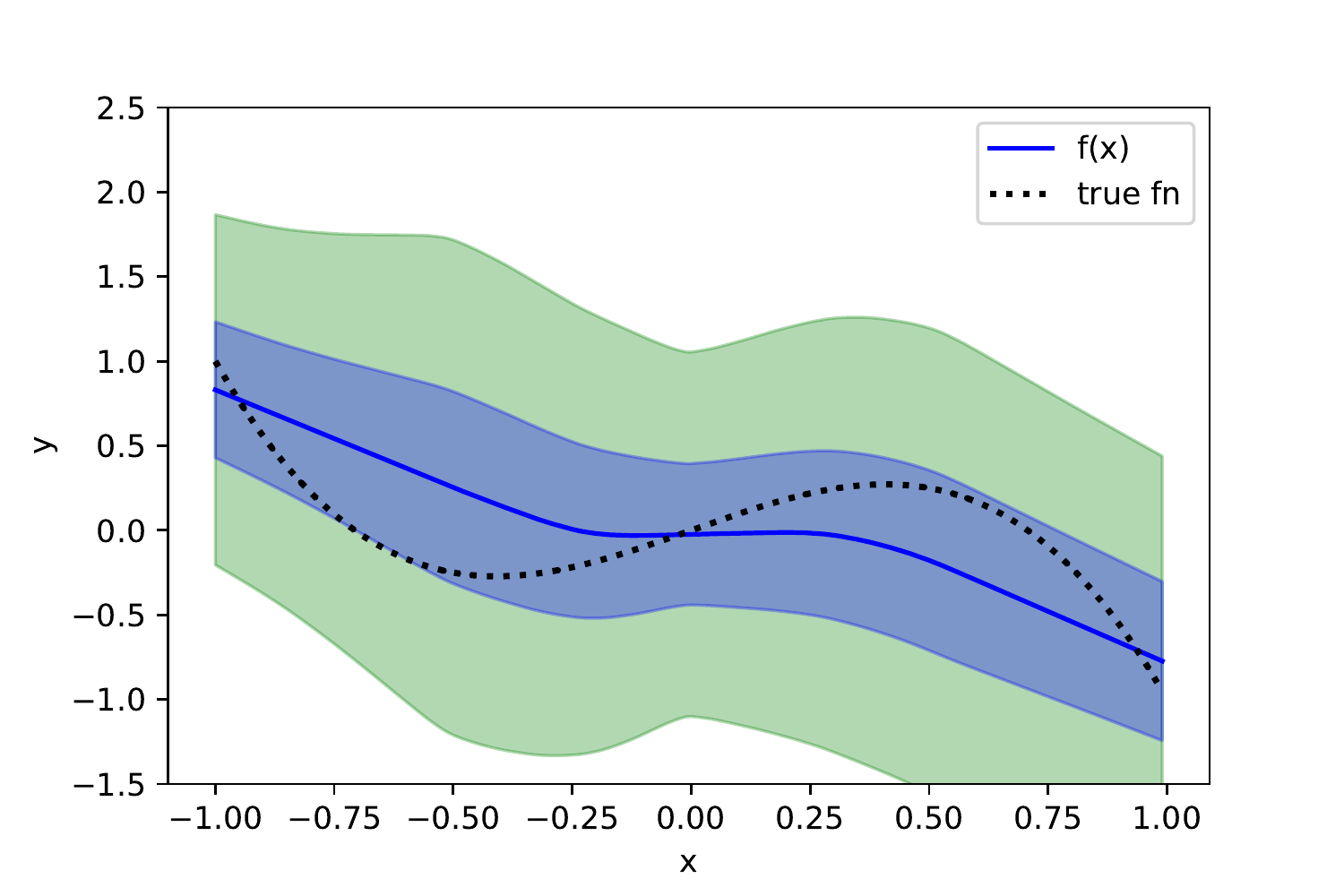}
\caption{Shaded areas show the output of 
a model trained to predict the RMSE of $f$ 
and 2.58 times the RMSE of $f$.}
\labelfig{compare_rmse}
\end{subfigure}
\hfill
\begin{subfigure}[b]{.49\linewidth}
\includegraphics[width=\linewidth]{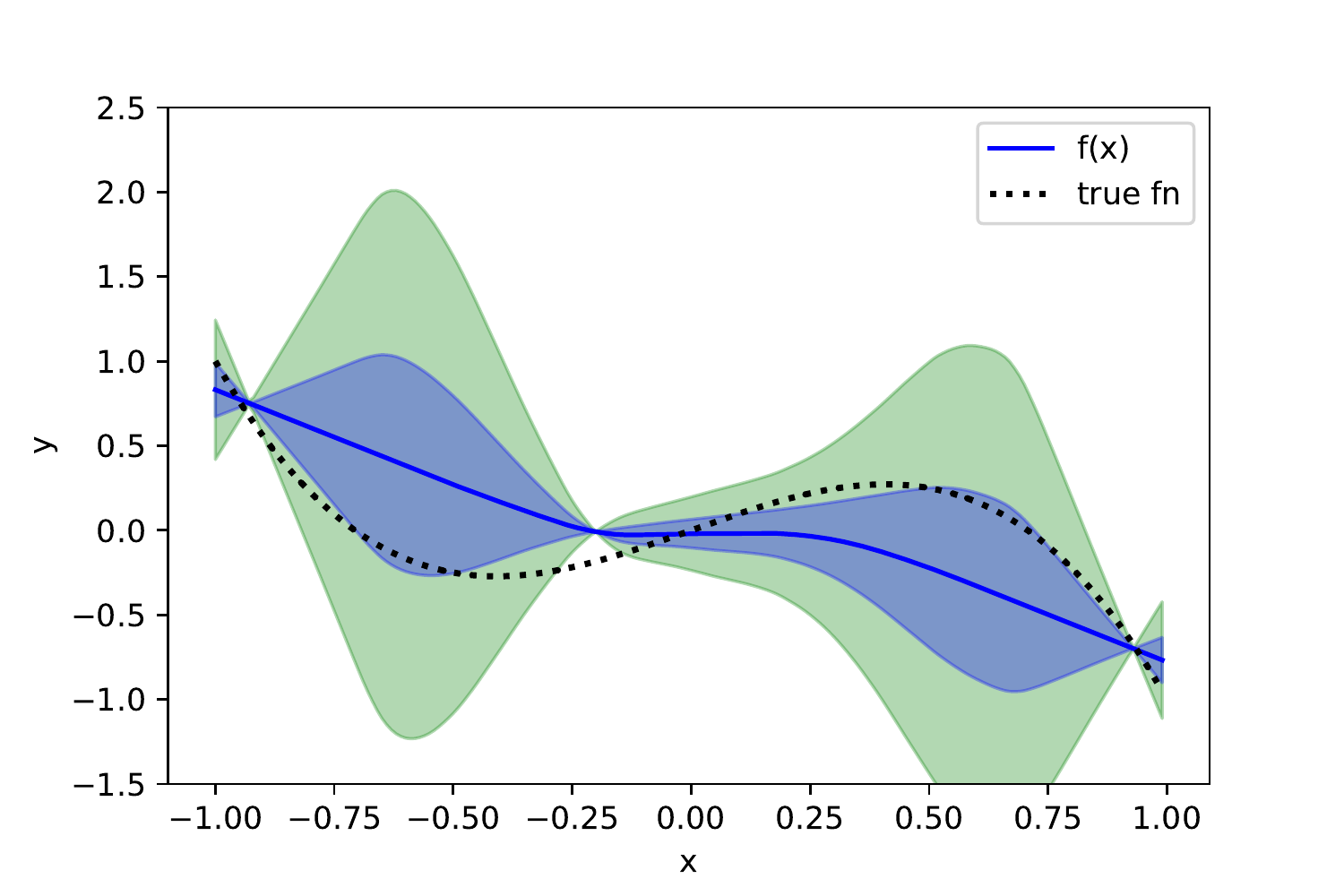}
\caption{Shaded areas show the residual overfit 
and 2.58 times the residual overfit.\\}
\labelfig{compare_rome}
\end{subfigure}

\caption{Using root mean squared error (RMSE) as the training target 
results in smoother interpolation of error bars 
but does not encourage exploration 
toward regions without support. 
In this example, 100 data points are generated 
at each point $x \in \{-1, -\frac{1}{2}, 0, \frac{1}{2}, 1\}$. 
In \reffig{compare_rmse}, 
the predicted uncertainty remains constant 
even though there is higher certainty 
at the observed inputs. 
In \reffig{compare_rome}, 
the highest regions of predicted uncertainty 
fall outside the observed inputs. 
Data are omitted for legibility. 
}
\labelfig{rmse_eb_toy}
\end{figure}

\section{Residual Overfit Method of Exploration (ROME)}

\todo{establish discrete actions and continuous context setting}

The properties of the squared residual overfit in \eq{residual_overfit_summary} 
are suggestive of 
the variance term in standard approaches 
for exploration-exploitation. 
Consider the discrete action set $a \in \mathcal{A}$ with context $x \in \mathcal{X}$ 
comprising continuous and/or discrete features. 
Before each interaction with the environment, 
the bandit observes a context $x$ 
and must score each action $a$. 
The action with the maximum score is taken greedily. 
The residual overfit may be applied 
to this setting using the following scores,
\begin{flalign}
	\hat{y}_a &= f(a, x) + \alpha | (f(a, x) - g(a, x) | \;\;\mathrm{(UCB)} \labeleq{ucb}\\
	\hat{y}_a &\sim p(f(a, x), (f(a, x) - g(a, x))^2) \;\;\mathrm{(Thompson\; sampling)} \labeleq{rome_ts}
\end{flalign} 
for some exploration hyperparameter $\alpha$. 
Exploration is guided towards actions 
where either the reward or error of $f$ is high, or the variance of $g$ is high. 
This approach is called the \emph{residual overfit method of exploration} (\name). 
Note that \namep uses a single sample (i.e. a dataset) Monte Carlo estimate of the 
expectation presented in \eq{residual_overfit_summary}. 
If pure exploration is required, the $f(a, x)$ value outside of the residual overfit terms in 
\eq{ucb} and \eq{rome_ts} may be replaced with 0.

\subsection{Exponential Family Moment Matching}

We consider how to apply the method to exponential families. 
Many distributions of interest belong to the exponential family, e.g., univariate or multivariate Gaussian, Bernoulli, multinomial, Poisson \citep{wainwright2008graphical}. 
In deep neural networks and decision trees, 
the last layer typically includes a member of the exponential family 
that induces a loss with respect to the observations and latent representation. 

Exponential family distributions over random variable $y$ take the form,
\begin{flalign}
	p(y \g \eta) &\propto \exp\left\{ t(y)^\top \eta \right\},
\end{flalign}
with natural parameter $\eta$ and sufficient statistics $t(y)$. 

Distributions in the exponential family 
have the property that 
the distribution that minimizes the KL divergence from 
the target distribution with a given mean and variance 
is obtained by matching moments. 
Here, we use $(f(x), (f(x) - g(x))^2)$ as the mean and variance.

\subsubsection{Binary Outcomes}

Use $\Beta(a, b)$ distribution to model the probability of binary outcomes. 
If $f$ has been trained to minimize Bernoulli (logistic) loss then its output is a suitable candidate 
as the mean of the Beta distribution, given its conjugate relationship with the Bernoulli 
(i.e. updating a Beta prior with Bernoulli evidence with Bayes' rule yields another Beta distribution). 
We can calculate the Beta pseudo-counts $(a, b)$ by matching the mean and variance,
\begin{flalign}
	\E[p_f] &= \frac{a}{a + b} \defeq f(x) \nnnl
	 \V[p_f] &= \frac{a b}{(a + b)^2 (a + b + 1)} \defeq (f(x) - g(x))^2 \\
	\implies a 	&= f(x) \left| \frac{f(x)(1-f(x))}{(f(x) - g(x))^2} - 1 \right| \nnnl
	b &= (1 - f(x)) \left| \frac{f(x)(1-f(x))}{(f(x) - g(x))^2} - 1 \right| \labeleq{rome_ts_beta}
\end{flalign}

If we interpret the output of $f$ as the Bernoulli probability $p_f$ 
of binary outcome $y$ with $\Beta$ prior, 
then the posterior parameters are equivalent to,
\begin{flalign}
	a &= \E[p_f] \left| \frac{\V_{p_f}[Y]}{\V[p_f]} - 1 \right| \nnnl
	b &= (1 - \E[p_f]) \left| \frac{\V_{p_f}[Y]}{\V[p_f]} - 1 \right|.
\end{flalign}

%\section{Broader Perspectives on \name}
\section{Bayesian Information Theoretical Perspective}
\labelsec{bayesian_method}

The analysis so far has required the errors of $f$ and $g$ to be independent, 
necessitating random data splits. 
It is not ideal to split the data because it will lead to worse estimates, 
especially with small data sizes. 
Is it possible to analyze \namep when $f$ and $g$ 
are not trained on different splits of the training data? 
To address this question and provide 
a broader perspective on the residual overfit, 
we now consider it in the setting of Bayesian information theory.

\labelsec{info_gain}

There are various information theoretic criteria for 
choosing the next action with Bayesian inference. 
A prominent class of methods maximizes the information gain 
for the parameters~$\theta$ from unknown response $y$ 
given dataset $\D$, 
\begin{flalign}
	x^* &= \arg_{x} \max \IG_{x}[\theta, y \g \D] \nnnl
	\mathrm{where\;\;} &\IG_{x}[\theta, y \g \D] = \Hh[ \theta \g \D] - \Hh_{x}[\theta \g \D, y] ,\labeleq{info_gain}
\end{flalign} 
where the subscript $x$ is used to indicate an arbitrary fixed query point. 
\eq{info_gain} is equivalent to maximizing the decrease 
in posterior entropy after the new observation \citep{mackay1992information}. 
Due to symmetry of information gain, 
\eq{info_gain} can be expressed 
in terms of entropy of the predicted target, avoiding unnecessary posterior updates~\citep{lindley1956measure, houlsby2011bayesian},
\begin{flalign}
	\IG_{x}[\theta, y \g \D]  &= \Hh_{x}[ y \g \D ] - \Hh_{x}[y \g \D, \theta] \labeleq{info_gain_t}
\end{flalign}
The model appears twice in \eq{info_gain_t}: 
in the first term with parameters marginalized out 
and in the second term with posterior averaged entropy. 
Due to this, 
the approach 
is unreliable when the predictive uncertainty of $y$ 
is underestimated, 
since it determines both the entropy and conditional entropy. 
In fact, in the limiting case of a point estimate for 
the posterior, such as the maximum \emph{a posteriori} (MAP) estimate, 
the information gain is zero across all~$x$. 

Our diagnosis of the problem of approximating information gain 
is that it is misleading to use the predictive entropy of $y$ 
under the \emph{typical} parameter 
in the first term of \eq{info_gain_t} (e.g. using variational inference or MAP). 
The peril is that the uncertainty of some actions is underestimated 
and leads to their elimination in an interactive setting, a type II error 
in identifying actions requiring exploration. 
In contrast, overestimating uncertainty (type I error) 
leads to self-correction over time 
as more samples are gathered. 
To reflect this explorative asymmetry, 
we consider the \emph{lowest upper bound} of the predictive entropy 
induced by an approximating distribution $q$. 
To formalize this notion, 
for any $q$, 
add a non-negative slack (recall, any KL divergence is non-negative) to the 
information gain, 
\begin{flalign}
	\IG_{x}[\theta, y \g \D] 
	&\le \hat{\IG}_{q,x}[\theta, y \g \D] \nnnl
	&:= \IG_{x}[\theta, y \g \D] + \KL{p(y \g \D, x)}{q(y \g x)} \nnnl 
	&= -\E_{p(y \g \D, x)}[\log q(y \g x)] - \Hh_{\D,x}[y \g \theta], \labeleq{info_gain_slack} 
\end{flalign}
then look for $q$ 
that minimizes $\hat{\IG}_{q,x}[ y, \theta \g \D]$ 
under the empirical average of $x \sim \D$. 
This is equivalent to minimizing $\E_{x \sim \D}\left[\KL{p(y \g \D, x)}{q(y \g x)}\right]$ 
which goes to zero as $q$ gets closer 
to the marginal predictive probability. 
\eq{info_gain_slack}~can be rewritten as the KL divergence 
between the prediction of the model and the approximation, 
\begin{flalign}
	\hat{\IG}_{q,x}[\theta, y \g \D]  &= \E_{p(\theta \g \D)}\KL{p(y \g \theta, x)}{q(y \g x)} \labeleq{kl_rome}.
\end{flalign}

\eq{kl_rome} may be estimated as follows. 
The approximation for the distribution of the outer expectation 
must be close to the true posterior $p(\theta \g \D)$. 
Hence, it is amenable to established methods for approximate Bayesian inference 
such as MCMC, variational inference, or MAP. 
As discussed before, 
the empirical $q$ approximation for the RHS of the KL divergence in \eq{kl_rome} 
minimizes $\E_{x \sim \D}\left[\KL{p(y \g \D, x)}{q(y \g x)}\right]$. 
When $p(y \g \D, x)$ is the true data generating distribution 
then $q$ is the maximum likelihood solution.

For a univariate Gaussian distribution on $y$, \eq{kl_rome} reduces to a closed-form expression,
\begin{flalign}
	\hat{\IG}_{q, x}[\theta, y \g \D]  &= \frac{1}{2 \sigma^2} \left(f(x) - g(x) \right)^2 + \mathrm{constant},
\end{flalign}
where $f(x)$ is the mean prediction of a MAP inferred model, 
$g(x)$ is the mean prediction of the $q$~model, 
and $\sigma^2$ is the irreducible variance. 
This recovers \namep in the deterministic exploration setting when $\sigma^2 = \frac{1}{2}$. 

\eq{kl_rome} extends to other observation likelihoods, e.g., the Bernoulli observation likelihood yields,
\begin{flalign}
	\hat{\IG}_{q,x}[\theta, y \g \D]  &=  g(x) \log \frac{g(x)}{f(x)} + (1-g(x)) \log \frac{1 - g(x)}{1 - f(x)},
\end{flalign} 
where $f(x)$ and $g(x)$ are the MAP and $q$ model probabilities of success, respectively. 
Poisson observation likelihood results in,
\begin{flalign}
	\hat{\IG}_{q,x}[\theta, y \g \D]  &=  g(x) \log \frac{g(x)}{f(x)} + f(x) - g(x),
\end{flalign}
where $f(x)$ is the mean predicted rate of the MAP model and $g(x)$ is the mean predicted rate of the $q$~model.

\subsection{Practicality of ROME}

When deploying a model in practice, 
it is standard procedure 
to perform a hyperparameter search 
to find the model architecture and training settings 
that perform best on held-out validation data. 
In this way, an overfit model, which we call $g$, 
is a by-product in the search for the best $f$. 
There are two main benefits to this observation. 
First, in this setting, \namep avoids additional training time. 
Second, there is low marginal engineering cost 
for deploying $g$, 
since it was at one point a candidate for $f$, 
so likely has much in common with $f$ such as 
features, targets, training algorithm, and deployment pipeline. 

What procedure should be used to select $g$ 
out of a candidate set of models? 
While it is likely advantageous to 
combine more than one overfit model, 
perhaps by cycling through them 
by iteration to encourage diversity,  
we have focused here on the case where 
there is a single $g$ for generality and ease of exposition. 
Theory suggests that 
$g$ should be selected to give 
the lowest variance unbiased estimate of the reward.

\cmt{online updating vs. retraining from scratch}

\section{Related Work}

There are various scalable algorithms 
that combine uncertainty quantification and model expressibility. 
Stochastic gradient Langevin dynamics (SGLD)~\citep{welling2011bayesian} adds noise to stochastic gradients 
in order to sample from a posterior distribution under appropriate conditions 
for the optimization surface and step size. 
SGLD applies to gradient-based models 
and, in a bandit setting, requires taking multiple gradient steps at prediction time 
to avoid correlated samples. 
Variational dropout~\citep{kingma2015variational, gal2016dropout} 
adapts the method of dropout regularization to perform 
variational approximation. 
It applies to deep neural networks and, 
since it is based on VI, 
underestimates posterior variance \citep{mackay1992information}. 
Bootstrapped Thompson sampling~\citep{osband2016deep} 
treats a set of models trained on bootstrap resamples of 
a dataset as samples of the parameters 
over which Thompson sampling may be applied. 
It can be used with the widest range of models 
but requires either 
resampling then training a model on each step 
or training multiple models from resamples in batch mode. 
If the rewards are sparse 
then a large number of resamples are required in batch mode. 

In practice, 
methods that perform shallow exploration are popular 
due to the ease with which they equip 
existing tuned models with exploration. 
Epsilon-greedy, Boltzmann exploration \citep{bianchi2017boltzmann}, and last-layer variance~\citep{snoek2015scalable} 
admit highly expressive models but ignore the uncertainty of most or all of the parameters \citep{riquelme2018deep}. 
This inflexibility may be compensated in some cases by an accurate tuning and decay of 
the exploration rate or temperature.

\section{Empirical Evaluation}

\cmt{Empirical questions:}

\cmt{What kind of overfitting works best in practice?}

\cmt{\namep failure modes?}

\cmt{In what settings does \name\, shine?}

In the empirical evaluation we 
compare \namep against several benchmarks 
and find that it performs competitively against 
both shallow and deep exploration methods. 
\later{
This supports the hypothesis that 
the gain from targeting exploration using the full model 
is greater than the 
approximation error highlighted in \refsec{residual_overfit_1} and \refsec{bayesian_method}. 
In more detail, there are two kinds of error that \namep introduces, 
bounding error and Monte Carlo approximation.  
Both vanish as the data size grows. 
Thus, the empirical question addressed here is 
whether these errors dominate 
in small data settings under cold start conditions. 
}

\paragraph{Methods}

The following methods are compared 
in the empirical evaluation, 

\begin{itemize}
	\item {\bf \name-TS}: \namep with Thompson sampling. Sample score from $\mathrm{Beta}(a,b)$ with pseudo-counts given by \eq{rome_ts_beta}. 
	\item {\bf \name-UCB}: \namep with upper confidence bound. Upper confidence bound score of $\mathrm{Beta}(a,b)$ with pseudo-counts given by \eq{rome_ts_beta}. 
	\item {\bf LinUCB}: contextual bandits with linear payoffs using the upper confidence bound~\citep{chu2011contextual}. 
	\item {\bf Epsilon greedy}: pick the action with the highest predicted reward with probability $(1-\epsilon)$ and a uniform random action with $\epsilon$ probability on each step. 
	\item {\bf Bootstrap-TS}: bootstrap the replay buffer and train $M$ models on each of the $M$ replications of the data~\citep{osband2016deep}. Thompson sampling is implemented by sampling one model uniformly each step and using its predicted rewards greedily to pick the action. 
	\item {\bf Uniform random}: pick the action uniform randomly on each step. Equivalent to epsilon greedy with $\epsilon = 1$. 
\end{itemize}

In the experiments, Bootstrap-TS uses $M=20$ replications, making training 20 times as computationally intensive as epsilon greedy and 10 times as computationally intensive as $\name$. 
For the UCB methods, the weighting for the upper bound is set to $\alpha = 1$ 
and $\epsilon = 0.1$ in epsilon greedy.

\todo{every 100 actions retrain}

With the exception of epsilon greedy and uniform random, all methods are
controlled with the same implementation settings using,
\begin{itemize} 
	\item a random forrest reward classifier model with the default settings from the \texttt{scikit-learn} package\footnote{version 0.21.1 https://scikit-learn.org/0.21/} 
	which uses an ensemble of 10 decision trees.
	\item A constant explore rate of 0.01 to mimic a small number of organic observations  
	arriving outside of the bandit channel \citep{sakhi2020blob}. 
	\item The model is retrained every 100 iterations. In real-world settings it is usually infeasible 
	to retrain after every interaction, necessitating batched interactions. 
\end{itemize}

\begin{table*}[t]
  \caption{Average regret with 95\% confidence interval over 10 replications.  
  Top performing method for each dataset in bold. 
  }
  \label{tab:results}
  \begin{tabular}{rccc}
    \toprule
    Method & \specialcell{Covertype\\(7 classes)} & \specialcell{Bach Chorales\\(65 classes)} &  \specialcell{MovieLens-\texttt{depleting}\\(3,600 classes)}\\
    \midrule
    LinUCB & $0.415 \pm 0.003$ & $0.664 \pm 0.007$ & $0.967 \pm 0.005$ \\  
    Epsilon Greedy & $0.403 \pm 0.005$ & $0.711 \pm 0.051$ & $0.970 \pm 0.000$ \\
    Bootstrap-TS & $\boldsymbol{0.390 \pm 0.003}$ & $0.668 \pm 0.028$ & $0.971 \pm 0.000$ \\
    \name-UCB & ${0.422 \pm 0.004}$ & $0.718 \pm 0.035$ & ${0.963 \pm 0.005}$ \\ 
    \name-TS & $0.524 \pm 0.004$ & $\boldsymbol{0.657 \pm 0.012}$ & $\boldsymbol{0.941 \pm 0.006}$ \\
    Uniform Random & $0.859 \pm 0.004$ & $0.985 \pm 0.001 $ & $0.986 \pm 0.001$ \\
    \bottomrule
  \end{tabular}
\end{table*}

\begin{figure}
\centering

\begin{subfigure}[b]{.31\linewidth}
\includegraphics[width=\linewidth]{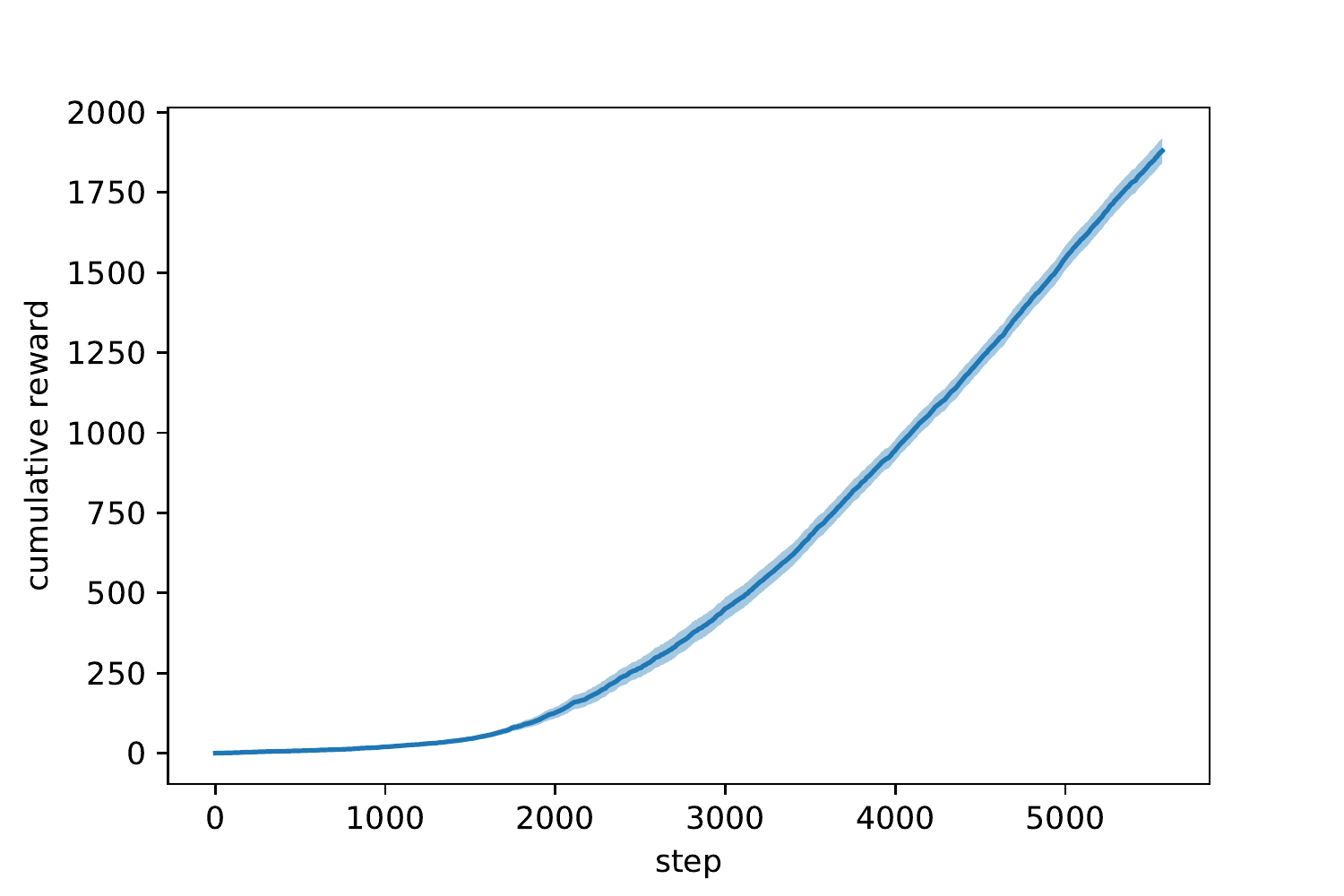}
\caption{\footnotesize LinUCB}
\end{subfigure}
\begin{subfigure}[b]{.31\linewidth}
\includegraphics[width=\linewidth]{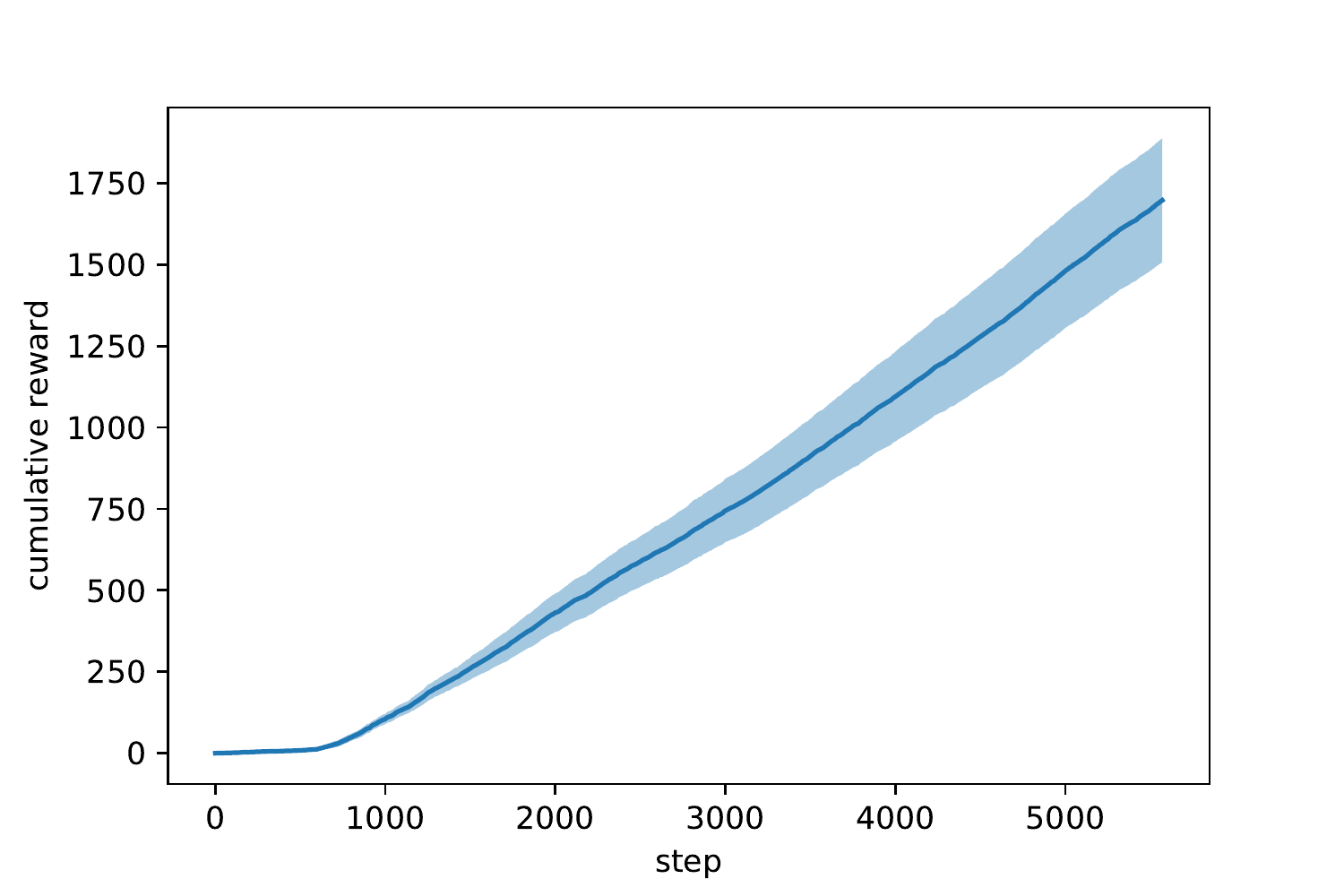}
\caption{\footnotesize Epsilon Greedy}
\end{subfigure}
\begin{subfigure}[b]{.31\linewidth}
\includegraphics[width=\linewidth]{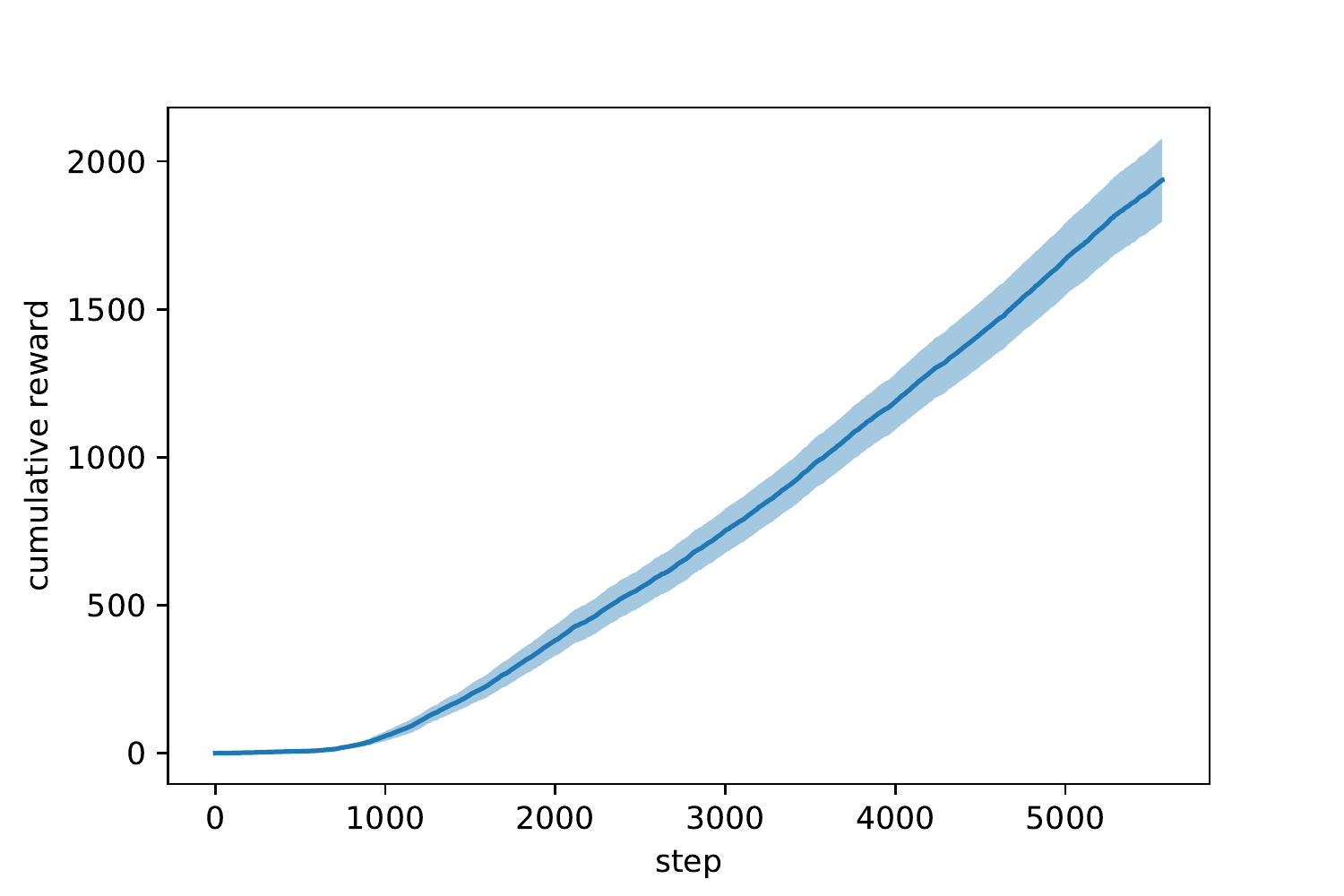}
\caption{\footnotesize Bootstrap-TS}\label{fig:gull}
\end{subfigure}

\begin{subfigure}[b]{.31\linewidth}
\includegraphics[width=\linewidth]{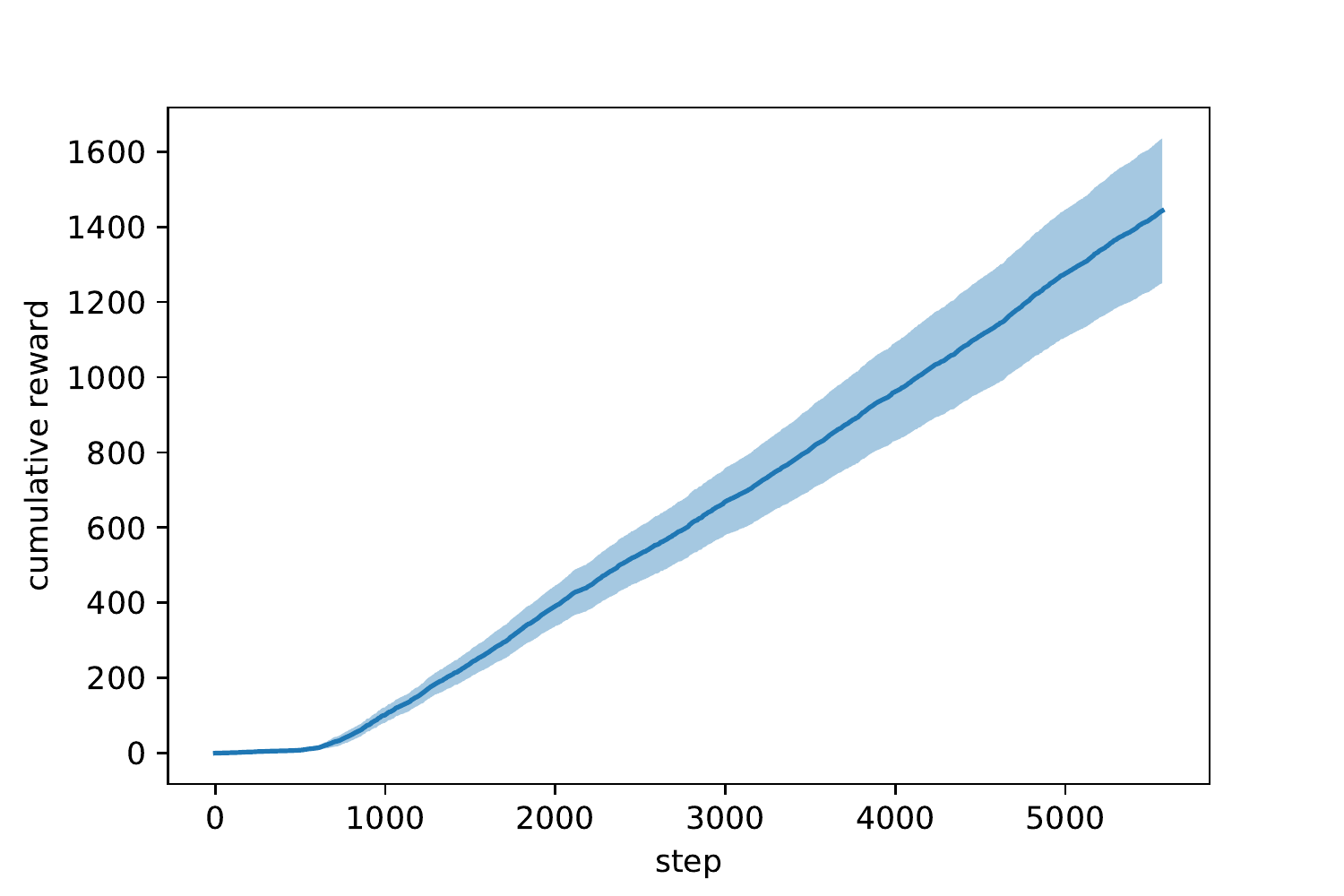}
\caption{\footnotesize \name-UCB}\label{fig:gull}
\end{subfigure}
\begin{subfigure}[b]{.31\linewidth}
\includegraphics[width=\linewidth]{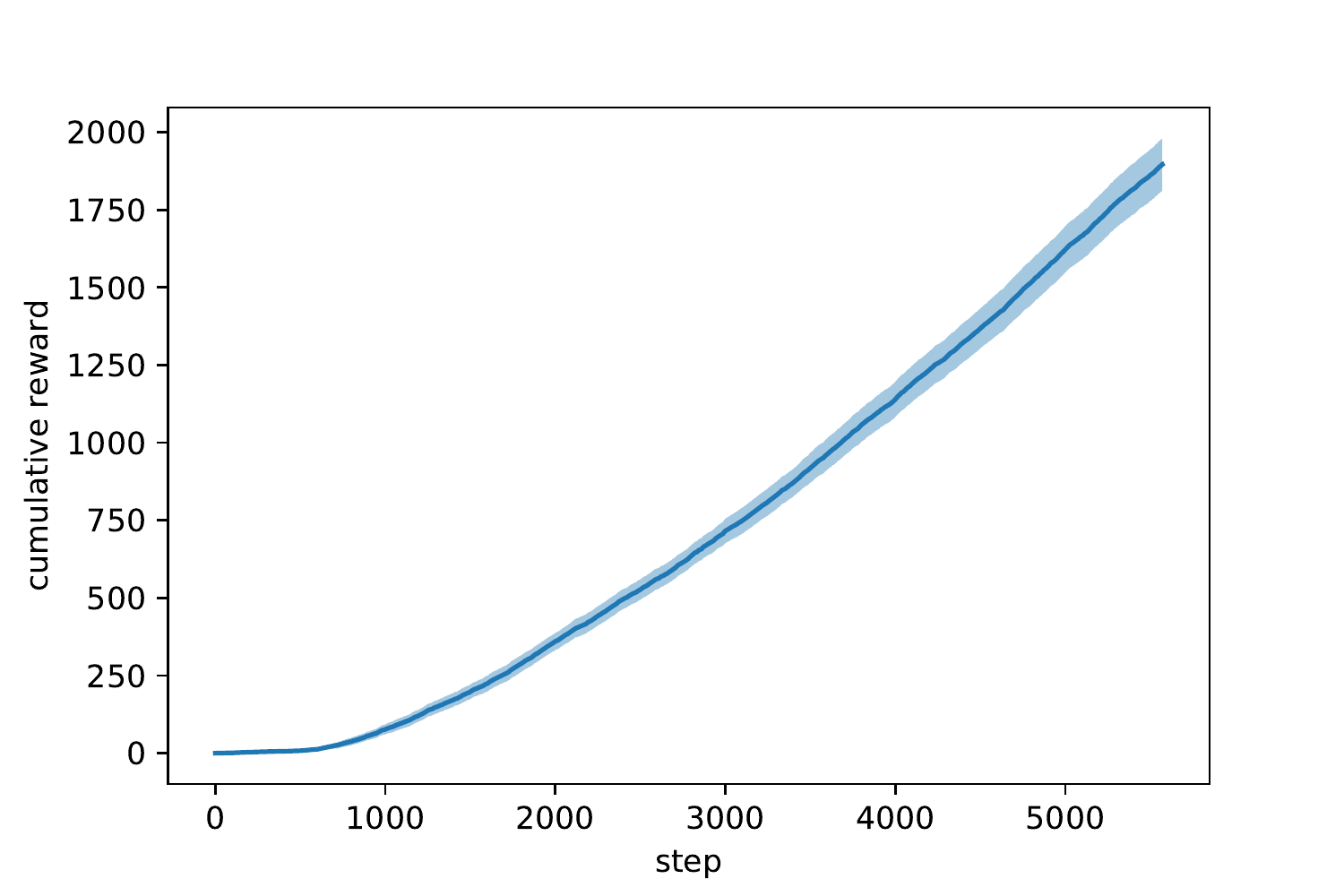}
\caption{\footnotesize \name-TS}\label{fig:gull}
\end{subfigure}
\begin{subfigure}[b]{.31\linewidth}
\includegraphics[width=\linewidth]{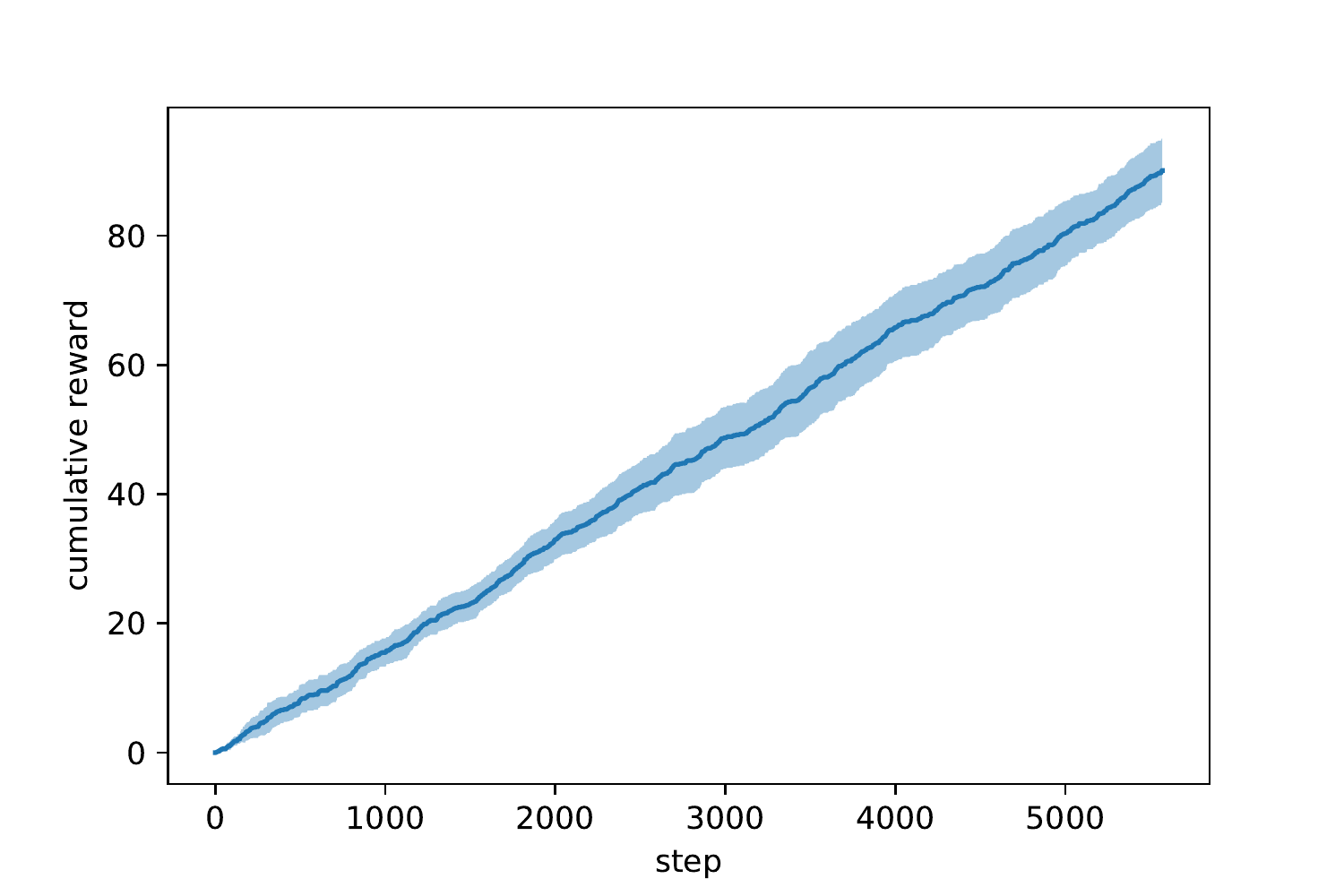}
\caption{\footnotesize Uniform Random}\label{fig:gull}
\end{subfigure}

\caption{Cumulative reward curves for the Bach Chorales dataset.}
\labelfig{bach_reward}
\end{figure}

\begin{figure}
\centering

\begin{subfigure}[b]{.31\linewidth}
\includegraphics[width=\linewidth]{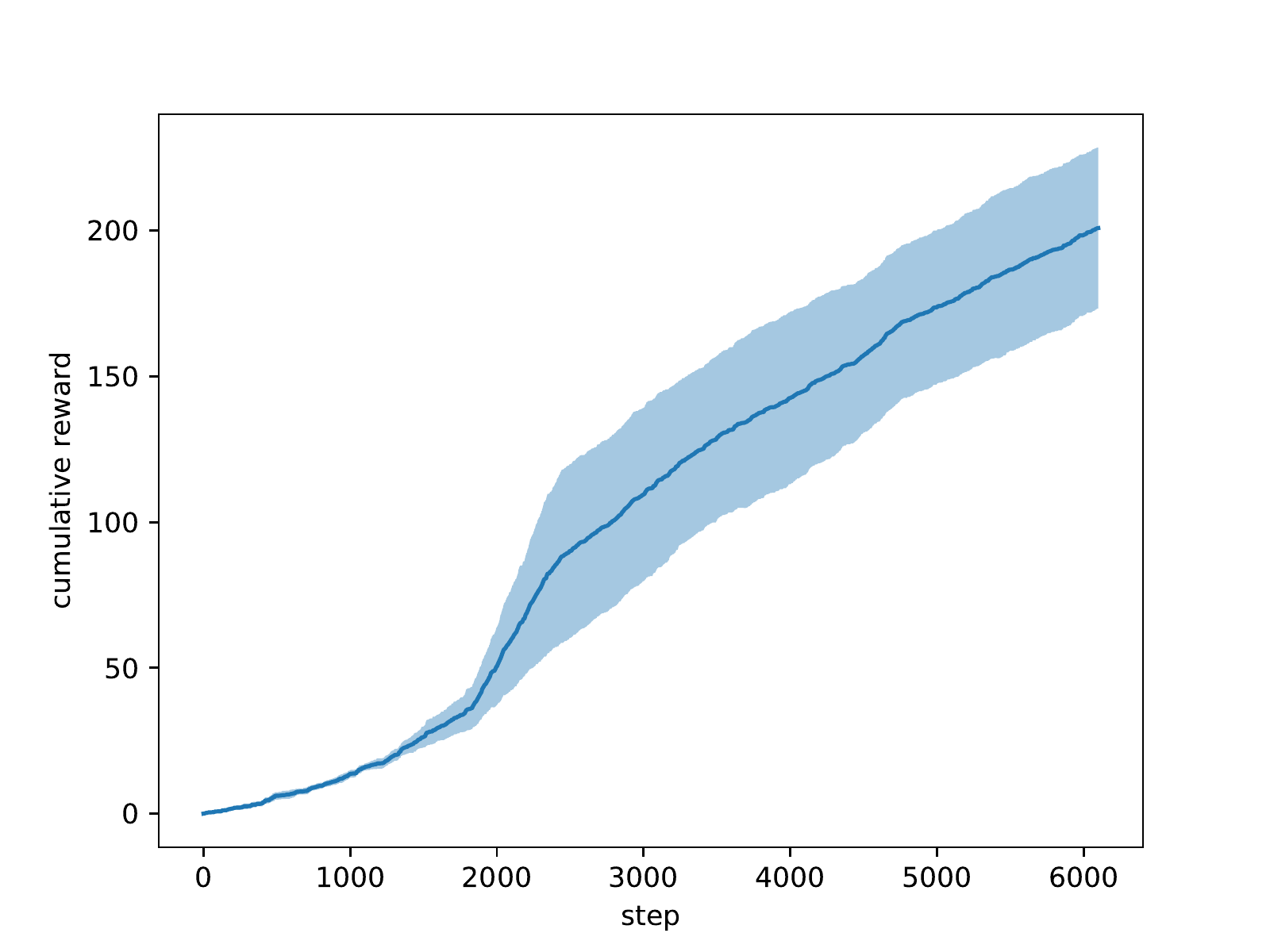}
\caption{\footnotesize LinUCB}
\end{subfigure}
\begin{subfigure}[b]{.31\linewidth}
\includegraphics[width=\linewidth]{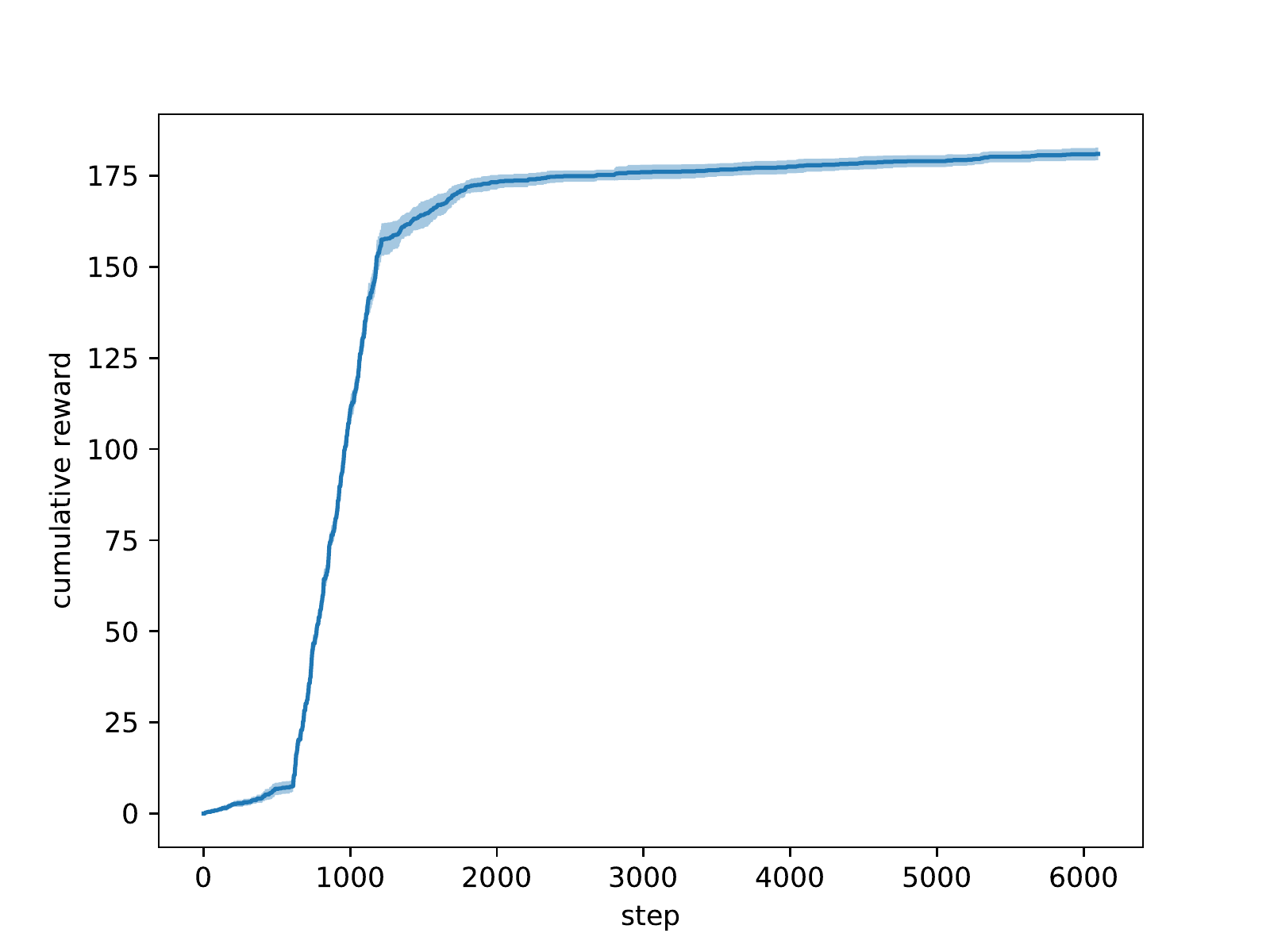}
\caption{\footnotesize Epsilon Greedy}
\end{subfigure}
\begin{subfigure}[b]{.31\linewidth}
\includegraphics[width=\linewidth]{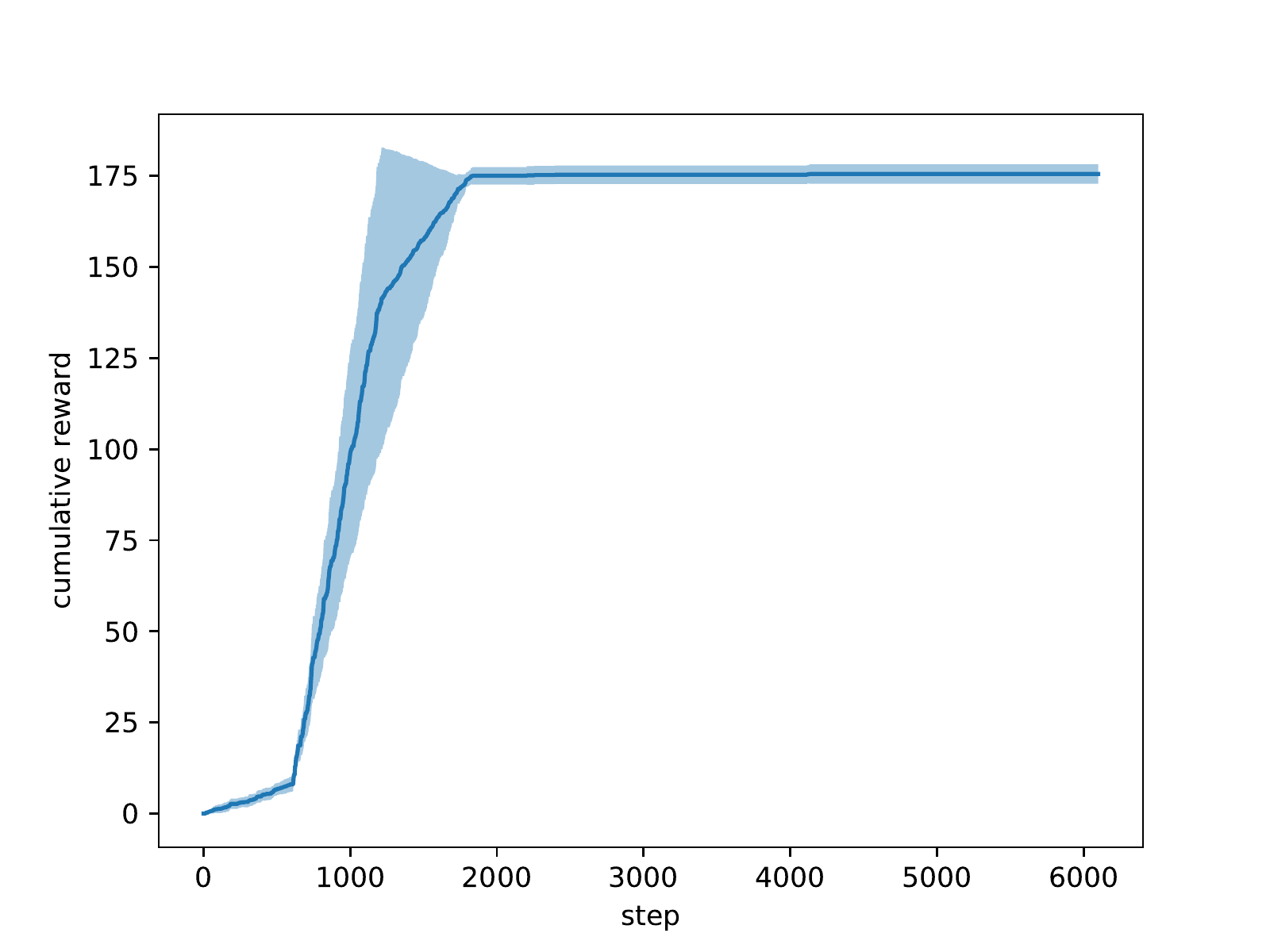}
\caption{\footnotesize Bootstrap-TS}\label{fig:gull}
\end{subfigure}

\begin{subfigure}[b]{.31\linewidth}
\includegraphics[width=\linewidth]{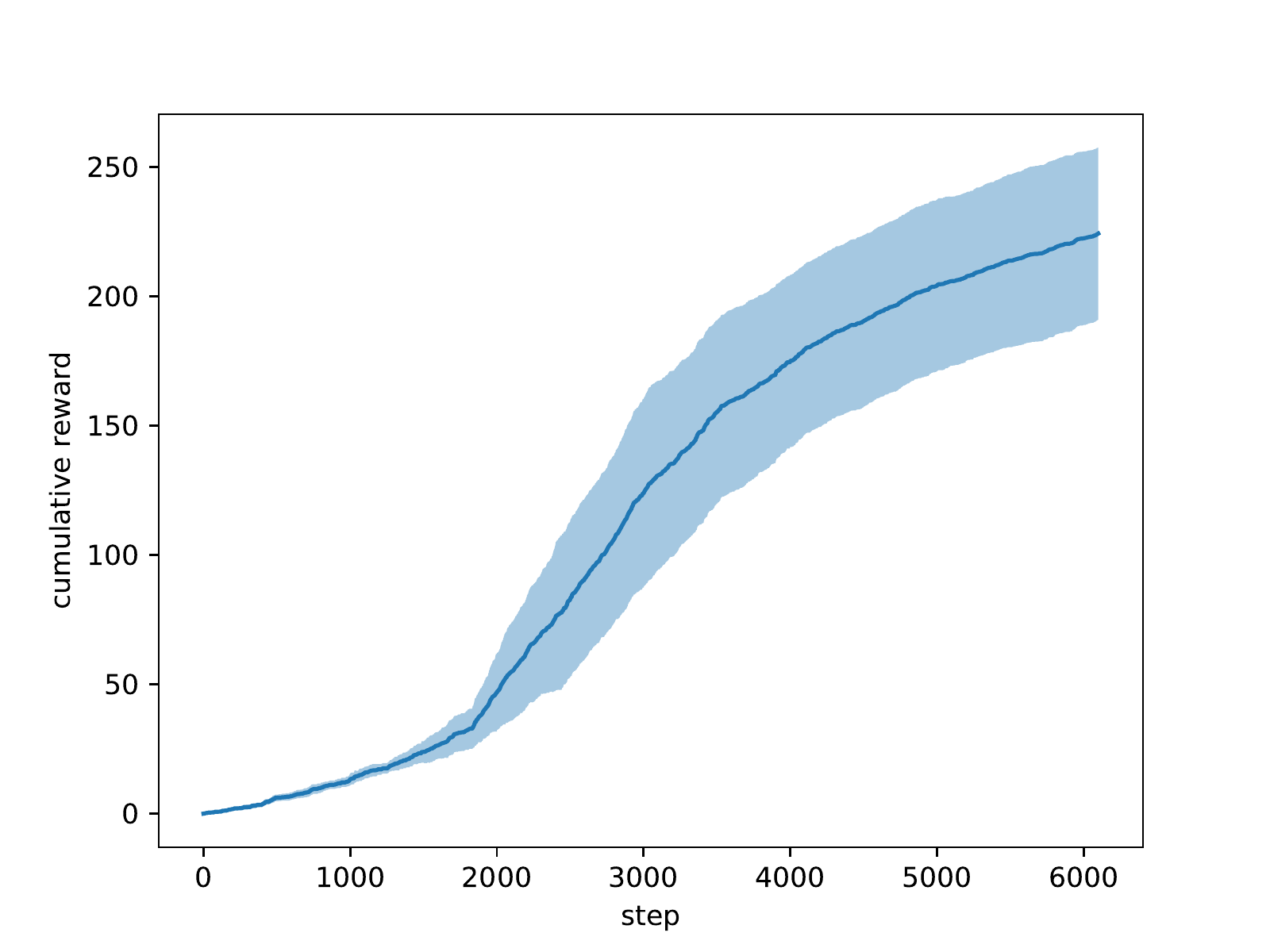}
\caption{\footnotesize \name-UCB}\label{fig:gull}
\end{subfigure}
\begin{subfigure}[b]{.31\linewidth}
\includegraphics[width=\linewidth]{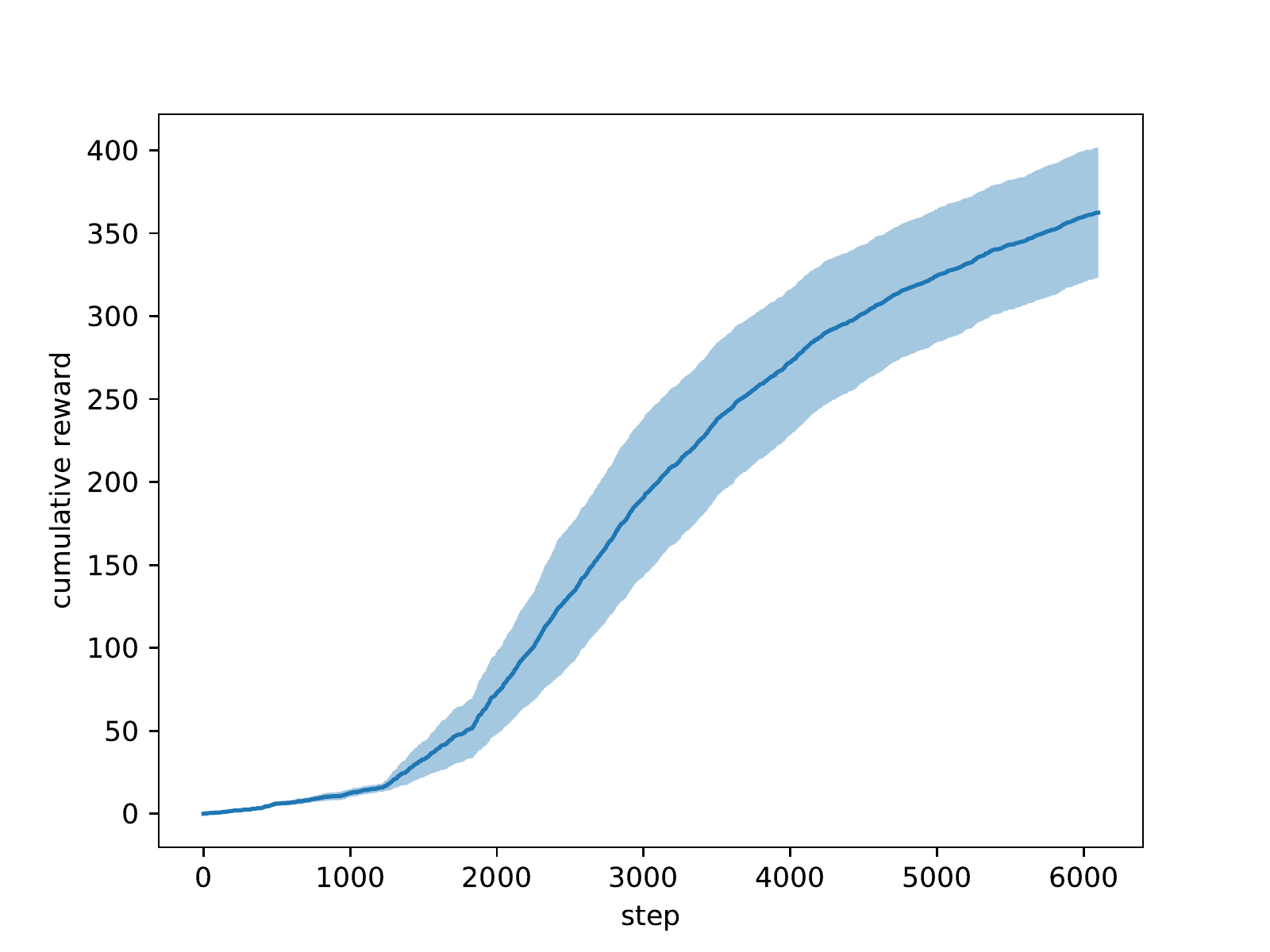}
\caption{\footnotesize \name-TS}\label{fig:gull}
\end{subfigure}
\begin{subfigure}[b]{.31\linewidth}
\includegraphics[width=\linewidth]{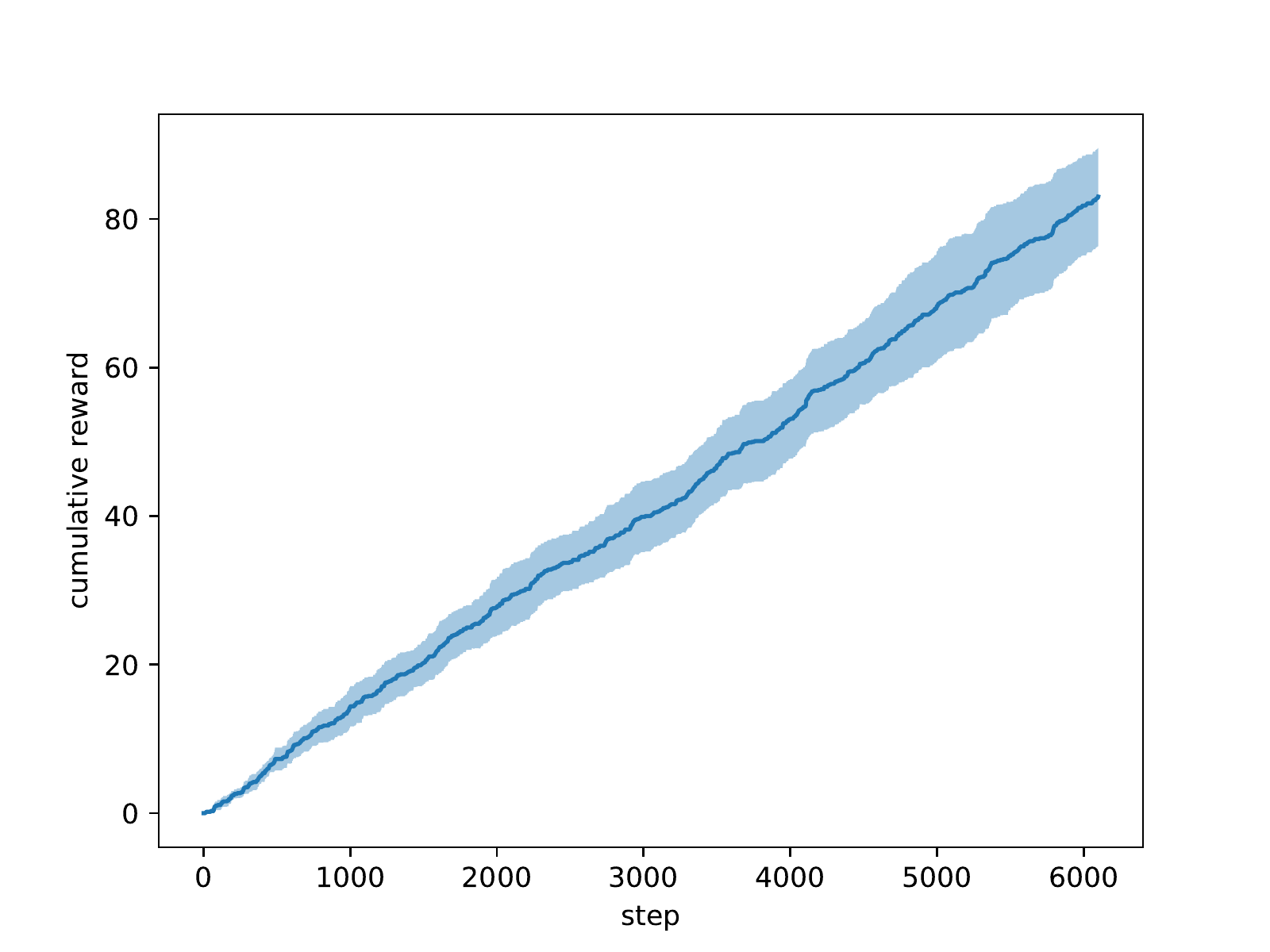}
\caption{\footnotesize Uniform Random}\label{fig:gull}
\end{subfigure}

\caption{Cumulative reward curves for the MovieLens-\texttt{depleting} dataset.}
\labelfig{mlens_reward}
\end{figure}

\paragraph{Datasets}

Evaluating explore-exploit performance 
requires setting up an interactive environment 
to assess the impact of acquired data on subsequent performance. 
We consider three classification datasets, 
and in each case, 
partial feedback is simulated by 
the environment providing a reward of 1 if 
the action corresponding true class is chosen 
and 0 otherwise. 
The instances are held in an arbitrary random order across 10 repetitions of the experiments. 
Actions are performed uniformly at random until every action has been observed at least once. 
The random seed for the methods and initial exploration varies across repetitions. 
The following datasets are studied,
\begin{itemize}
	\item {\bf Covertype}: data comprises 7 classes of forest cover type predicted from 54 attributes over 581,012 instances.\footnote{https://archive.ics.uci.edu/ml/datasets/covertype}
	\item {\bf Bach Chorales}: Bach chorale harmony dataset with 17 features to predict 65 classes of harmonic structure in 5,665 examples.\footnote{https://archive.ics.uci.edu/ml/datasets/Bach+Choral+Harmony}
	\item {\bf MovieLens-\texttt{depleting}}: a matrix of 100,000 interactions between 610 users and 7,200 items \citep{harper2015movielens}. 
	To replicate the cold start task recommender systems face 
	when introducing new items, 
	we split the items randomly into two equal-sized groups: 
	existing items and cold start items. 
	The bandit has access to all the historical user and existing item interactions 
	but the interactions between users and cold start items receives only partial feedback. 
	In each step of the experiment, the bandit chooses which 
	cold start item to recommend to a user based on the observed 
	interaction and context history. 
	The bandit makes 10 passes through the dataset. 
	To replicate the depleting effect of consuming items, 
	the same cold start item and user may give a reward of 1 no more than once, 
	and subsequently 0. 
\end{itemize}

\todo{add a text dataset? e.g. reuters}

\paragraph{Results} 
In Table~\ref{tab:results}, we find that \namep performs well across the datasets. 
Bootstrap-TS performs best for small action spaces 
where the rewards are denser. 
As the number of actions grows, 
it becomes harder for a small number of positive examples 
to appear in a significant number of bootstrap samples. 
Across datasets, the Thompson sampling approaches (\name-TS and Bootstrap-TS) 
outperformed the UCB methods. 
This is likely due to the benefit of stochasticity 
in the batch action setting 
in addition to the strong empirical performance for Thompson sampling 
observed more generally \cite{chapelle2011empirical}. 

\reffig{bach_reward} shows 
the cumulative reward curves of the methods 
as a function of the number of interactions 
in the Bach Chorales dataset. 
Early on, 
LinUCB explores more than the other model-based approaches, 
and as a result, 
Bootstrap-TS achieves the highest cumulative reward at the end. 
The most challenging dataset was MovieLens-\texttt{depleting}, 
due to both the large action space (3,600 actions) 
and depleting rewards. 
For this dataset, \reffig{mlens_reward} illustrates how only LinUCB, \name-TS, and \name-UCB were able 
to continue discovering high value actions after 10 passes through the dataset.\footnote{This holds for Uniform trivially since it depletes the popular items much slower than the other methods.}

%\todo{ \section{Conclusions \& Future Work} }

\section{Conclusions}

In this paper we developed theoretical and empirical justifications 
for the merits of combining a tuned and overfit model 
for exploration. 
The residual overfit method of exploration (\name) 
%combines a biased and unbiased model to identify 
approximately identifies actions and contexts with the highest parameter variance. 
The method can be applied to explore-exploit settings 
by adding the best regularized estimate of the reward. 
We provided a frequentist interpretation 
and a Bayesian information theoretic interpretation 
that shows that the residual overfit approximates an upper bound 
on the information gain of the parameters. 
Experiments comparing \namep with widely used alternatives 
shows that it performs well at balancing exploration and exploitation.

\begin{acks}
We thank Nikos Vlassis, Ehsan Saberian, Dawen Liang, Pannaga Shivaswamy, Maria Dimakopoulou, Yves Raimond, Dar\'io Garc\'ia-Garc\'ia, and Justin Basilico for their insightful feedback.
\end{acks}

\bibliographystyle{plainnat}
\bibliography{rome}

\appendix

\end{document}